\documentclass[journal]{IEEEtran}

\usepackage{cite}
\ifCLASSINFOpdf
   \usepackage[pdftex]{graphicx}
\usepackage{epstopdf}
\else
\usepackage[dvips]{graphicx}
\usepackage{epstopdf}
\fi

\usepackage{array}
\ifCLASSOPTIONcompsoc
  \usepackage[caption=false,font=normalsize,labelfont=sf,textfont=sf]{subfig}
\else
  \usepackage[caption=false,font=footnotesize]{subfig}
\fi

\ifCLASSOPTIONcaptionsoff
  \usepackage[nomarkers]{endfloat}
 \let\MYoriglatexcaption\caption
 \renewcommand{\caption}[2][\relax]{\MYoriglatexcaption[#2]{#2}}
\fi
\usepackage{url}
\usepackage{color}
\usepackage{cite}

\usepackage{setspace}
\usepackage{tikz}
\usetikzlibrary{shapes.geometric, arrows}
\begin{document}
\title{Drowsiness Detection Based On Driver Temporal Behavior Using a New Developed Dataset}
\author{\IEEEauthorblockN{F. Faraji, F. Lotfi, J. Khorramdel, A. Najafi, A. Ghaffari}\\
	\IEEEauthorblockA{K. N. Toosi University of Technology, Tehran, Iran.\\
	Email: farnooshfaraji@email.kntu.ac.ir}
}
%\author{Faraz~Lotfi, Saeideh~Ziapour, farnoosh~Faraji and Hamid~D.~Taghirad,~\IEEEmembership{Senior Member,~IEEE}% <-this % stops a space
%\thanks{Authors are with the Advanced Robotics and Automated System (ARAS),
%Department of Systems and Control,
%Faculty of Electrical and Computer Engineering, K.N. Toosi
%University of Technology, Tehran, Iran, e-mail:
%f.lotfi@email.kntu.ac.ir, z.saideh@gmail.com, FarnooshFaraji@email.kntu.ac.ir, taghirad@kntu.ac.ir.}}


\maketitle

%\doublespacing
\begin{abstract}
Driver drowsiness detection has been the subject of many researches in the past few decades and various methods have been developed to detect it. In this study, as an image based approach with adequate accuracy, along with expedite process, we applied YOLOv3 (You Look Only Once-version3) CNN (Convolutional Neural Network) for extracting facial features automatically. Then, LSTM (Long-Short Term Memory) neural network is employed to learn driver temporal behaviors including yawning and blinking time period as well as sequence classification. To train YOLOv3, we utilized our collected dataset alongside transfer learning method. Moreover, the dataset for LSTM training process is produced by the mentioned CNN and is formatted as a two-dimensional sequence comprised of eye blinking and yawning time durations. The developed dataset considers both disturbances such as illumination and drivers head posture. To have real-time experiments a multi thread framework is developed to run both CNN and LSTM in parallel. Finally, results indicate the hybrid of CNN and LSTM ability in drowsiness detection and the effectiveness of the proposed method.
%%A comparison between the results of this paper with some recent works indicate that the proposed method is significantly more effective for detection of driver drowsiness. 
\end{abstract}

\begin{IEEEkeywords}
Drowsiness detection and prediction, Convolutional neural network, Long-Short Term Memory, Hybrid Approach 
\end{IEEEkeywords}

\section{Introduction}
\IEEEPARstart{D}{rowsiness} is a leading cause of accidents. As it is reported by National Highway Traffic Safety Administration(NHTSA), drowsy driving claimed more than 792 lives in 2019. Since monitoring the driver awareness is of high importance, %NREM\footnote{Non Rapid Eye Movement} is a stage of sleep process that can be distinguished from three stages of transplantation: from wake to sleep, light sleep and deep sleep and 
scientists are mostly focused on the first stage of sleep process which named drowsiness phase~\cite{ref1}. Primarily, the state of driver drowsiness is detected by monitoring three types of measurements: physiological-based (using EEG, ECG, EOG and EMG signals~\cite{ref2}), vehicle-based (focusing on deviations from lane position, movement of the steering wheel~\cite{ref3} and pressure on the acceleration pedal~\cite{ref4}) and behavioral-based measurements (including yawning, eye closure, head pose etc.~\cite{ref5}). %In first type; EEG is utilized along with ECG, EOG and EMG, and drowsiness detection is accurately obtained through sensor fusion~\cite{ref3}. Despite of precise results, the main drawback of this measurement is being invasive. New generation of sensors can be connected through Wi-Fi and Bluetooth, however; recipience noises make them unreliable~\cite{ref4}.The second type focused on deviations from lane position, movement of the steering wheel~\cite{ref5} and pressure on the acceleration pedal~\cite{ref6}. %In this method, many criteria including conditions of the roads and the accuracy of the sensors are highly influential. In third method the driver behaviors, including yawning, eye closure, head pose etc. are monitored~\cite{ref7}. 
According to the prior tests, best performance of drowsiness estimation is achieved by tracking eye behavior~\cite{ref6} hence, in this study eye state changes and driver yawning is selected to be focused. Existing methods for classification of open and closed eyes can be largely divided into two categories; 1) Non-image based such as electro-oculography (EOG), which captures the electrical signals of the muscles in vicinity of eyes and 2) Image-based methods. In the latter approaches, the first step is determining the eyes location and making decision on its closeness/openness. In this regard, there are various image processing algorithms for extracting features. For instance, there are methods based on support vector machine (SVM)~\cite{ref7}, neural networks~\cite{ref8} and probability density functions (PDF)~\cite{ref9} and histogram of oriented gradient-based SVM (HOG-SVM)~\cite{ref10}. Since it is difficult to find the optimal number of mentioned features in images, we proposed a CNN-based method, based on its classification capability. Multiple architectures provide the ability to recognize visual patterns directly from pixel images with minimal preprocessing. For instance, AlexNet, ZFNet, ResNet, VGGNet and YOLO are designed with different number of convolutional layers, pooling layers and fully connected layers for distinct applications~\cite{ref11}. As stated in~\cite{ref12}, ResNet50, VGG16 and InceptionV3 are trained, and the features are fused to detect facial movements such as blinking, yawning and swaying with $78\%$ accuracy. In~\cite{ref13} the Viola-jones Algorithm is utilized to detect face and eyes, then a four-layer-convolutional classifier is trained to detect closed eye as a symbol of drowsiness. In this method, the momentary behavior is considered and achieved $96.4\%$ with accuracy. Naurois et.all~\cite{ref14} applied MATLAB neural network toolbox to detect and predict the driver drowsiness. In this study, they achieved $4.18$min as a predicted time via using driver behavior, psychological and vehicle measurements simultaneously. Multi task convolutional neural networks (MTCNN) in accordance with facial landmarks are investigated to detect face in complex situations and determine the eye and mouth closure with the angle of the landmarks~\cite{ref15}. The experimental results show $92\%$ accuracy. In~\cite{ref16} a method is provided which uses VGG18 as a pretrained CNN along with a camera installed inside the car; as a result, it is claimed that the accuracy reaches $99\%$. This approach not only detects driver distraction but also the cause of that by analyzing the driver images. Sughra Razzaq et al.~\cite{ref17} investigated a hybrid method which tracks driver facial features consist of eyes and mouth closeness/openness and road lane departure; they applied skin segmentation to remove the unnecessary background from face images and MATLAB vision techniques to detect lane departures. Moreover, using combination of two cameras capturing the road and drivers video data, they predicted whether the driver is unconscious or not. Another kind of neural networks are recurrent neural networks(RNN) which use prior information to perform present task. They are mostly used in language modeling, however; video sequence detection and image captioning can be considered as well. Moreover due to their limited memory, LSTM networks are developed such that can select the information to remember or forget, and they are capable of learning long-term dependencies. 
%As stated in~\cite{ref15} AdaBoost algorithm is used to detect face region because of its robustness and after finding the eye pupil location, PCA (principal component analysis) is applied to recognize eye state. Then, calculating the eye closure time (PERCLOS) will determine whether the driver is drowsy or not. G. M. Bhandari et al. employed Haar features in~\cite{ref16} to find mouth location and detected driver yawning by mouth geometric as a sign of drowsiness. 
Weiwei Zhang and Jinya Su in~\cite{ref18} discussed driver yawning detection as a feature of being drowsy. In their system a CNN (Google Net) is trained to extract the features of images and an LSTM to analyze temporal characteristics which leads to accurate results. %In~\cite{ref20} both CNN and LSTM are used to improve innovative method for drowsiness detection. It is done in two main parts; spatial domain and temporal domain. In spatial part, the method tries to extract facial features such as eyes and the mouth in one frame and in temporal domain, LSTM is adopted to remember the data of the past frames. Since every frame transition does not give slight changes, developers improved time skip to ignore useless data and increase calculations speed. In~\cite{ref1}both CNN and facial landmarks are helped to detect drowsiness. First the face and eyes are detected via Viola and Jones algorithm then landmarks will be placed. Finally, by calculating the distance of eyes landmarks in different time scales, they could estimate the level of drowsiness in accordance with eye closure time. 
As it is mentioned, deciding on driver drowsiness level is mostly affected by eye features and the mouth plays a significant role as well. In this work, transfer learning approach is used to train the pretrained YOLOv3 CNN with our own dataset developed on face, opened/closed eyes, mouth/yawn and eyebrow detection.
This CNN is utilized to produce a two-dimensional sequence of closed eye and yawning duration time series that is used as LSTM dataset. Finally, with the help of LSTM, drowsiness detection besides prediction is realized.
The rest of this paper  as follows: Section II focuses on our own dataset, in section III the perposed method for drowsiness detection/prediction is presented IV experimental results are reported and finally in section V conclusions are articulated.

\section{Our Dataset}
\begin{figure}[t]
	\centering
	\hspace*{4ex}\subfloat[view.1]{\includegraphics[width=1in]{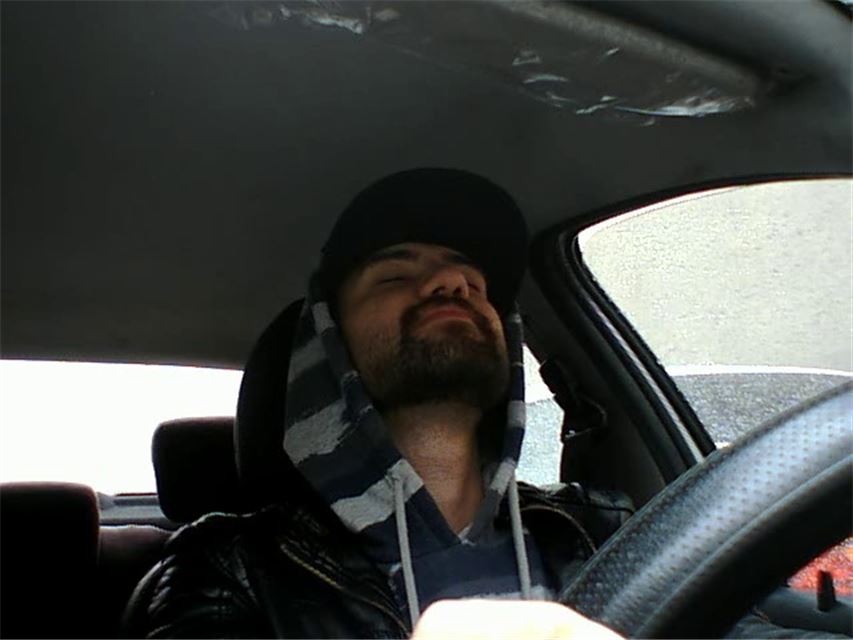}
	   	\label{a-phase}}
	\\
	\subfloat[view.2]{\includegraphics[width=1in]{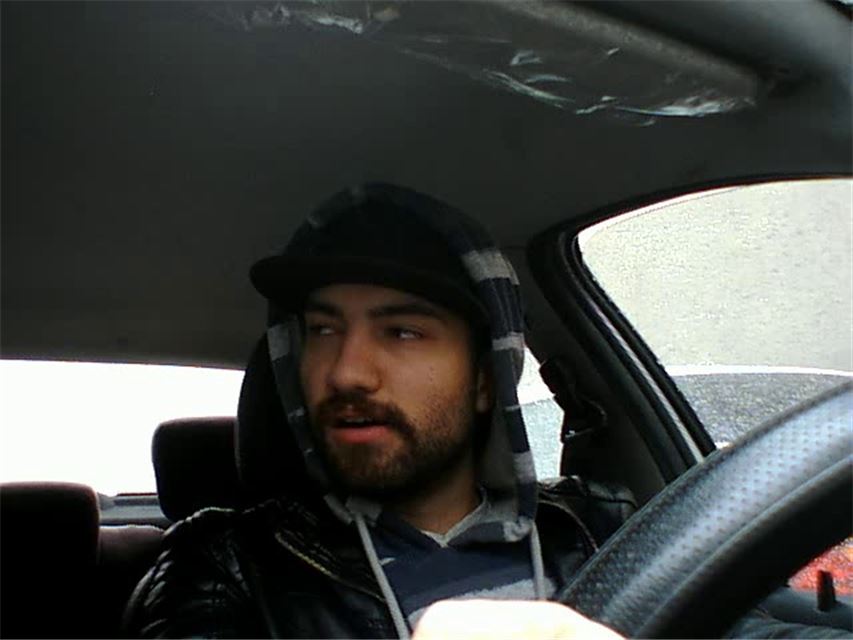}
		\label{b-phase}}
	\subfloat[view.3]{\includegraphics[width=1in]{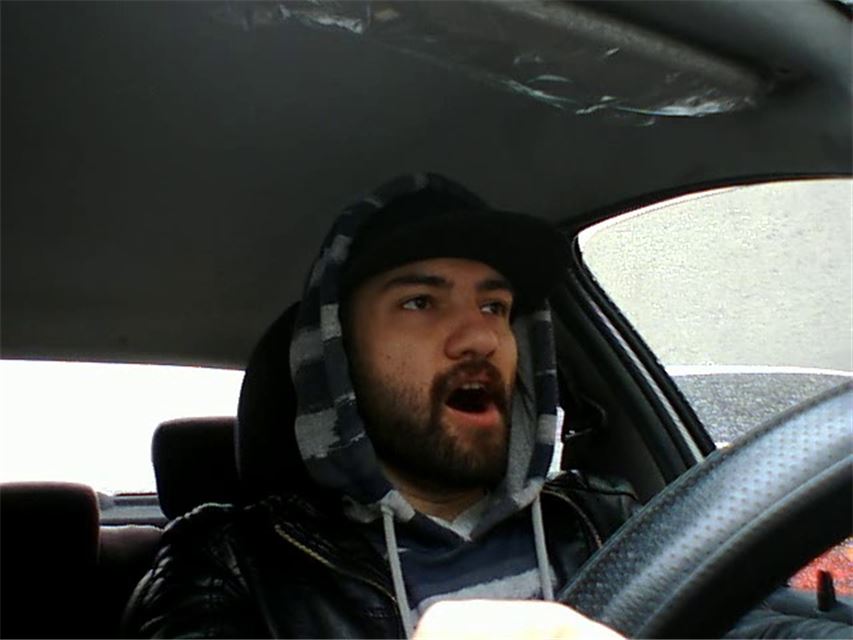}
		\label{c-phase}}
	\subfloat[view.4]{\includegraphics[width=1in]{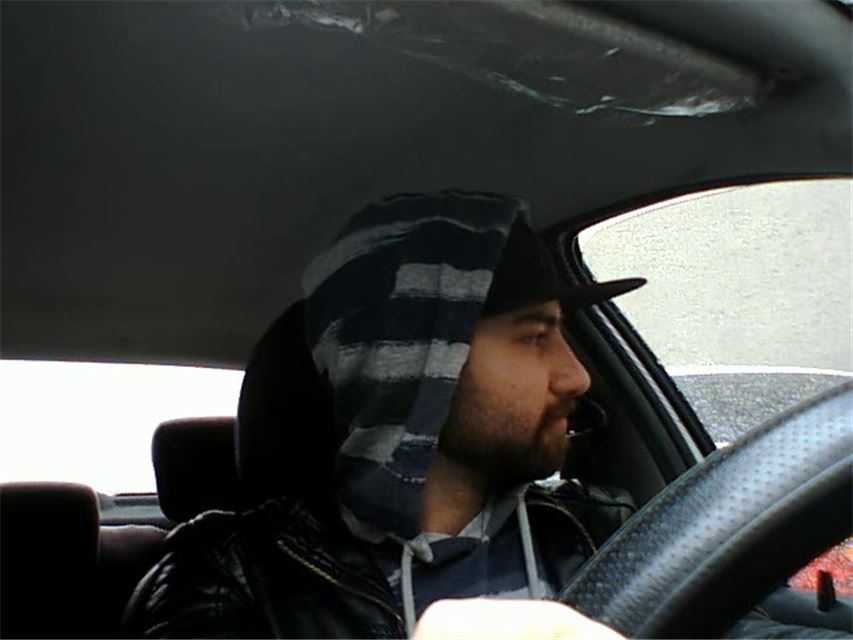}
		\label{d-phase}}
	\\
	\hspace*{4ex}\subfloat[view.5]{\includegraphics[width=1in]{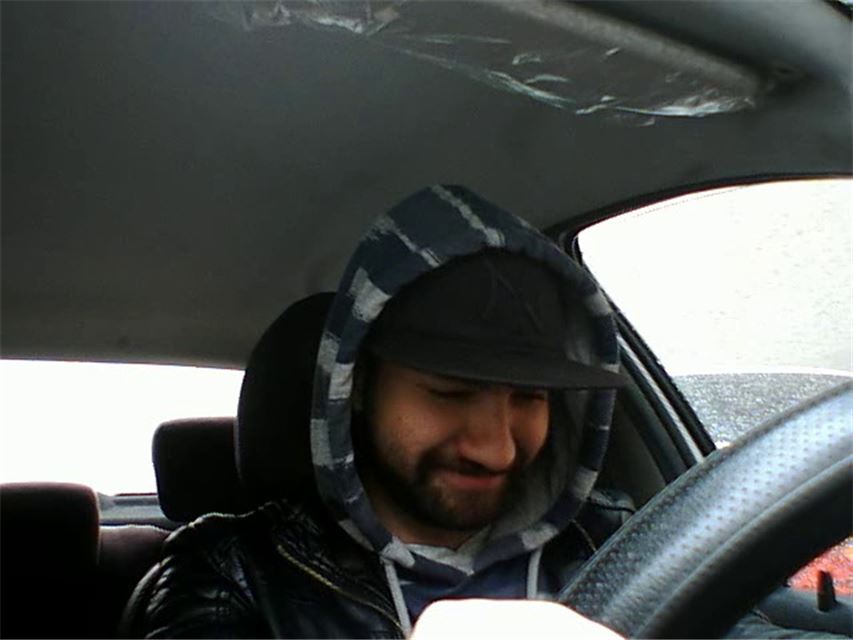}
		\label{e-phase}} \caption{Images with different head pose}
	\label{fig2}
\end{figure}

\begin{figure}[t]
	\centering
	\subfloat[image]{\includegraphics[width=1.7in]{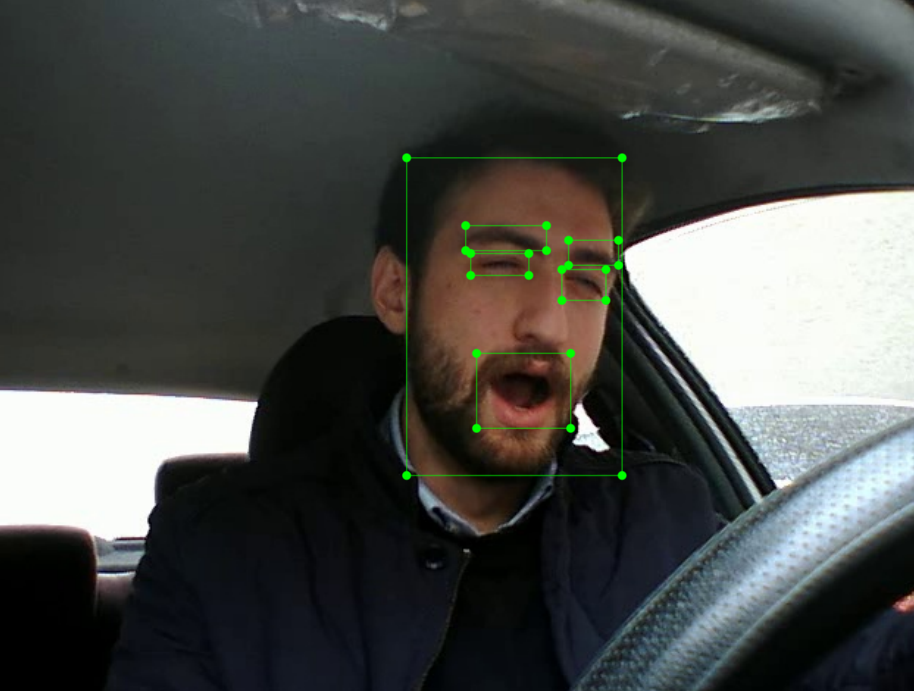}
		\label{fig5}}
	\hfil
	\subfloat[part of annotation]{\includegraphics[width=1.9in]{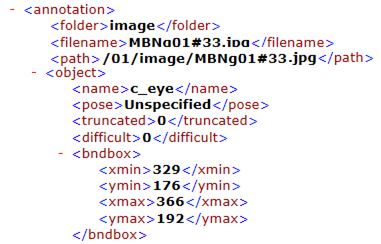}
		\label{fig6}} \caption{Labeled image and annotation file in dataset}
	\label{fig3}
\end{figure}

The invented dataset mainly has two parts. Part 1 is a group of labeled images for training CNN and part 2 is a developed time series for training LSTM network. Described methods in introduction section performed high accuracy rates. They, however, faced some limitations due to changing conditions including illumination, head pose, wearing glasses, facial expression etc. Although different datasets are made and available such as DDD~\cite{ref19} and YawDD~\cite{ref20}, we generated our own dataset which carefully consists of most conditions to train our neural network as accurate as possible and robust to the variations. To make dataset for part$1$, a $PS3$ camera with $640\times480$ pixels resolution is installed in the middle of dash and recorded $30fps$ RGB video of participants who sat in the car to have better simulation of driving mode. Mentioned procedures are done during the daylight for different lightening conditions and participants were asked to look at points $a$ to $e$ as depicted in Fig.~\ref{fig2} while they talk, laugh, look at sides, yawn and close their eyes smoothly to provide a rich dataset.
Finally, $28$ videos in $.avi$ format achieved from $18$ male and $10$ female candidates. and videos audio were removed in order to reduce the file size. 
Next, $1042$ number of images are extracted with enough variations in $.jpg$ format by a python code, 
from videos. This method of extracting images, prepares them as many as required, more or less than $1042$, based on neural network depth.
 The following abbreviations in Table \ref{table1}, have been used to name the participant folders, consist of images and videos. In this method of naming, information about images containing gender, light condition and wearing glasses or not are accessible. Additional information including the permission for publishing images in the paper are available in dataset documents. 
\begin{table}[h]
	\renewcommand {\arraystretch}{1.4}
	\caption{Abbreviations} \label{table1}
	\centering \begin{tabular}{ccc}
		\hline \hline \bfseries gender & \bfseries light & \bfseries glasses or not \\
		\hline F: female & B: bright & G: yes \\
		\hline M: male & D: dark & Ng: no \\ 
%		\hline  & S: shine &  \\		
	\end{tabular}
\end{table}\\
For instance, $FBNg21$ stands for a Female in Bright who does not wear glasses and she is the $21^{st}$ participant. Images are manually labeled. %with bounding boxes by means of python coding. 
As a result, a $.xml$ file is made as an annotation for each image that is required for training and validation process in supervised learning. Sample of a labeled image with bounding boxes and its annotation file is depicted in Fig. \ref{fig3}; as it is indicated, the annotation file consists of each object along with its correspondence bounding box location. 
Initially, eyebrows were not determined in our model. It turns out, a few head posture inducements CNN wrongly considers eyebrow as closed eyes (e.g. Fig. \ref{a-phase}). To resolve it, we add eyebrow to the labels. Labels consist of face, mouth, yawn, opened eye, closed eye and eyebrow. To enhance the results, the produced dataset is augmented with DDD dataset. In the first phase, selected CNN is trained on our dataset, then evaluated on DDD videos. For the second phase, 250 challenging images, consisting sunglasses and low light situation were extracted. the enhanced dataset is utilized for training and the results are reported.   
Second part of the dataset is needed for LSTM neural network. In order to train RNNs, it is essential to have past sequences information. For example, in language processing, a sequence of characters is utilized through the time.
As a closely related work in~\cite{ref13} open or closeness of eyes and yawn were saved as binary values in the feature vector in order to train an LSTM. In this work, to train an LSTM a dataset is developed that comprises the sleeping stages for $10$ people and unlike the mentioned method eye closure and yawning duration time are saved as two-dimensional time series data instead of binary values. The values are in millisecond and stored as a $.txt$ file to be used for LSTM training.

\emph{Remark:} The dataset for LSTM NN has 2D dimensions. One for blinking and the other for yawning.  
\section{Methodology}
This section is dedicated to our methodology to detect and predict driver drowsiness based on images which are recorded while driving. Considering real-time operation as a highly important issue in driver drowsiness detection, a fast and accurate neural network is needed. In this research YOLOv3 CNN is applied as a pretrained network, which is proved to be utilized as a powerful means for object detection \cite{ref21}. %As delineated in Fig. \ref{fig4} YOLO in given architecture consists of 53 convolutional layers called Darknet-53. %The output layer is changed to detect only 6 defined objects (as in Fig. \ref{fig8}) due to its application.

Since it is rare to have dataset with sufficient size for training CNN weights from scratch, transfer learning method have been used commonly.
%\begin{figure}[h]
%	\centering
%	\includegraphics[width=2.7in]{pics/Yolov31}
%	\caption{YOLOv3 architecture \cite{ref24}}
%	\label{fig4}
%\end{figure}
In this regard, there are three major scenarios in transfer learning as follows: employing CNN as a fixed feature extractor, fine-tuning CNN and deploying a pretrained model. As our dataset is partly small, the second scenario is selected. In fine-tuning method, the strategy is not only replacing and retraining the classifier on top layers with new dataset, but also to fine-tuning the weights of the pretrained network by continuing the backpropagation~\cite{ref22}. 
In order to follow this scenario, last weights of training YOLOv3 on COCO dataset \cite{ref22} is used as initial values in training procedure on our dataset. %During training the following multi-part cost function is optimized \cite{ref27}: 
%\begin{eqnarray}
%\label{equation1}
%&\lambda_{coord}\sum_{i=0}^{S^2}\sum_{j=0}^{B} \L_{ij}^{obj}[(\sqrt{w_i}-\sqrt{\hat{w_i}})^2+(\sqrt{h_i}-\sqrt{\hat{h_i}})^2]
%\nonumber \\
%&+\lambda_{coord}\sum_{i=0}^{S^2}\sum_{j=0}^{B} \L_{ij}^{obj}[(x_i-\hat{x_i})^2+(y_i-\hat{y_i})^2]
%\nonumber \\
%&+\sum_{i=0}^{S^2} \L_{i}^{obj}\sum_{c\in classes}(p_i(c)-\hat{p_i}(c))^2
%\nonumber \\
%&+\lambda_{noobj}\sum_{i=0}^{S^2}\sum_{j=0}^{B} \L_{ij}^{noobj}(c_i-\hat{c_i})^2
%\nonumber \\
%&+\sum_{i=0}^{S^2}\sum_{j=0}^{B} \L_{ij}^{obj}(c_i-\hat{c_i})^2
%\end{eqnarray}
%where the followings have to be mentioned: \\
%1- $\lambda_{noobj}=0.5$ and $\lambda_{coord}=5$. \\ 
%2- $\L_{ij}^{obj}=1$ if the $j^{th}$ bounding box in cell $i$ is responsible for detecting the object, otherwise 0.\\
%3- $\L_{i}^{obj}=1$ if an object appears in cell $i$, otherwise 0.\\
%4- $w_i$, $h_i$, $x_i$ and $y_i$ denote the $i^{th}$ bounding box $width$, $height$ and $upper-left$ $corner$ position, respectively.\\ 
%5- ${p_i}(c)$ denotes the conditional class probability for class $c$ in cell $i$ and ${c_i}$ is the box confidence score of the box $j$ in cell $i$. Furthermore, the parameters with " $\hat{ }$ " stand for network output while others show ground truths.
   
On the other hand, recurrent neural networks are efficient solutions for learning a scenario like sleeping. Although it is not easy to train RNNs due to gradient explode or vanish, LSTM as a special type of RNN is capable of learning long-term dependencies. %As it is shown in Fig. \ref{fig7} its structure consists of $3$ gates called: input, output and forget. %First, a $tanh$ function is used for input activation function where $W$ and $V$ are weights for the input and previous layer output and $b$ denotes the bias~\cite{ref17}.
%\begin{figure}[t]
%	\centering
%	\includegraphics[width=2.7in]{pics/LSTM1}
%	\caption{LSTM cell }
%	\label{fig7}
%\end{figure}
%\begin{eqnarray}
%\label{equation2}
%g=\tanh(x_{t}W^{g}+h_{t-1}V^{g}+b^{g})
%\end{eqnarray}
%Then a sigmoid activation function decides on how much information should go through the gate as an input, and the output of the input stage is multiplied element-wise by $i$ and $g$.
%\begin{eqnarray}
%\label{equation3}
%i=\sigma(x_{t}W^{i}+h_{t-1}V^{i}+b^{i})
%\end{eqnarray}
%The next stage is the forget gate where the internal state $s_{t}$ appears and delayed by one-time step. Afterwards, another sigmoid activation function operates on the values.
%\begin{eqnarray}
%\label{equation4}
%f=\sigma(x_{t}W^{f}+h_{t-1}V^{f}+b^{f})
%\end{eqnarray}
%The output of the forget stage is:
%\begin{eqnarray}
%\label{equation5}
%s_{t}=s_{t-1}\circ f+g\circ i
%\end{eqnarray}
%As depicted in Fig. \ref{fig7} the output gate and final output of the cell can be expressed as:
%\begin{eqnarray}
%\label{equation6}
%o=\sigma(x_{t}W^{o}+h_{t-1}V^{o}+b^{o})
%\end{eqnarray}
%\begin{eqnarray}
%\label{equation7}
%h_{t}=\tanh(s_{t})\circ o
%\end{eqnarray}
\begin{figure}[t]
	\centering
	\includegraphics[width=3.2in]{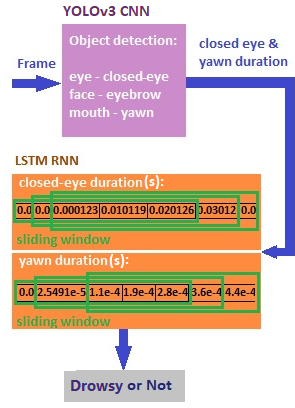}
	\caption{Schematic of the proposed method}
	\label{fig8}
\end{figure}

In this research an LSTM is used to do sequence classification. Additionally, the mentioned YOLO CNN is exerted to produce sequence of closed eye and yawning time with $0.1sec$ sample time. Then, a sliding window with $5sec$ size moves toward this sequence with determined stride that in this case is $1$ and the input for the LSTM is prepared. Since CNN has more complicated computations, multi thread coding is assigned to apply CNN and LSTM together in parallel. As shown in Fig. \ref{fig8}, each frame passes through CNN and LSTM, to estimate the probability of drowsiness and warn the driver. Note that, the driver drowsiness prediction can be realized via monitoring this value; in which additive variations in this value can inform the driver about the incoming drowsiness.
\section{Experimental results} 
\begin{figure}[t]
	\centering
	\includegraphics[width=2.5in]{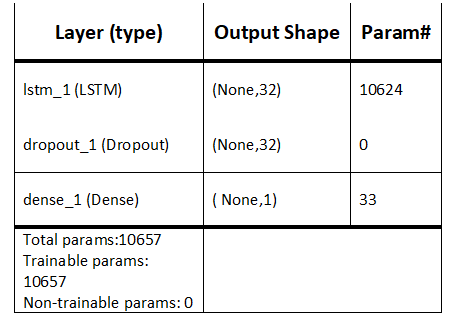}
	\caption{LSTM architecture }
	\label{fig11}
\end{figure} 
This part accentuated the results of implemented method to evaluate its efficiency. Both CNN and LSTM are employed to analyze yawn and eye closure and to detect the driver drowsiness. After the detection, eye closure and yawning duration are saved as time series data; then a predetermined sliding window with length of $5$ is passed through this data sequence and classification of the driver behaviors dynamic is done via utilizing this window as the LSTM input. Analyzing video frames takes more time than numerical data time series; additionally, the size of LSTM is less than the CNN in this paper, thus, CNN is slower than LSTM and if both work in succession in a same thread, some of the data will be lost. In order to solve this problem a multi thread algorithm is applied that helps CNN and LSTM work in parallel. In other words, after a while that the first LSTM input is prepared through CNN detections; at the time the LSTM is processing its input, the CNN is also processing the next image in parallel to produce the next sample for LSTM input sequence. This method minimizes the lost information in missed images and the total system speed is optimized. Furthermore, neural networks have been run on GPU to obtain sufficient speed of $30 fps$.
\begin{figure}
	\centering
	\subfloat[view.1]{\includegraphics[width=2.5in]{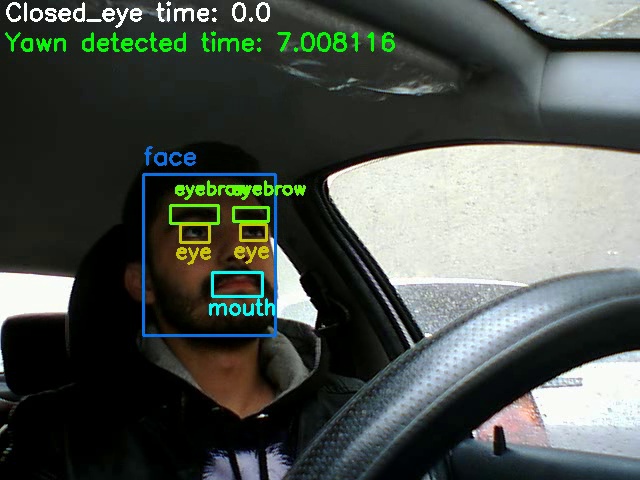}
		\label{a9}}
	\\
	\subfloat[view.2]{\includegraphics[width=2.5in]{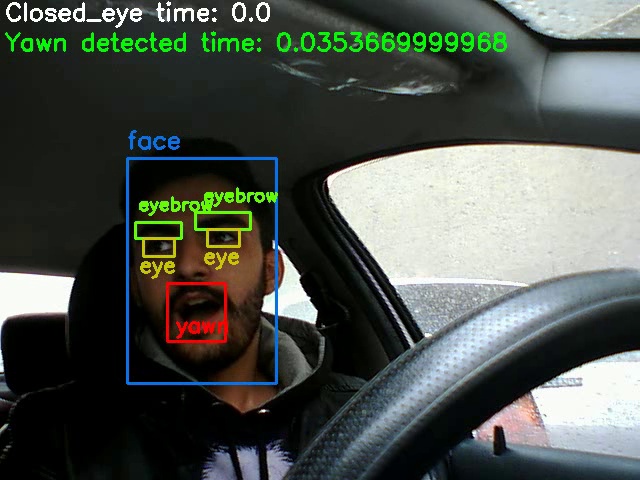}
		\label{c9}} 
	\\
	\subfloat[view.3]{\includegraphics[width=2.5in]{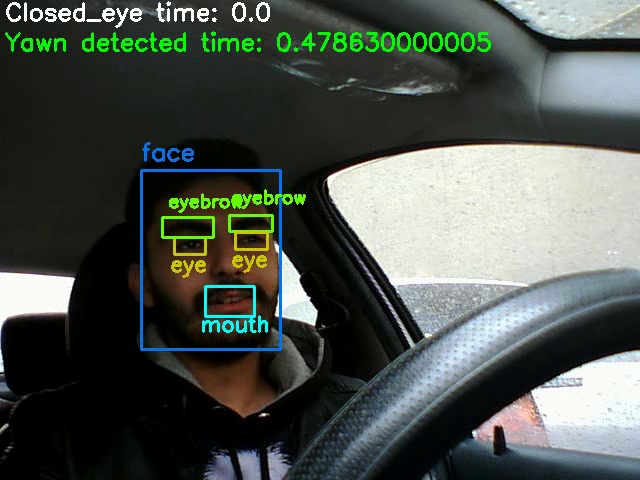}
		\label{e9}} \caption{CNN results for different head pose with no glasses. part I}
	\label{fig9_1}
\end{figure}
\begin{figure}
	\centering
	\subfloat[view.5]{\includegraphics[width=2.5in]{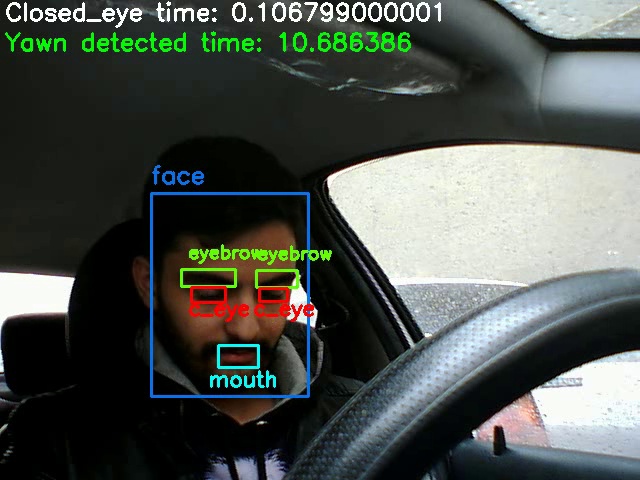}
		\label{d9}} \
	\\
	\subfloat[view.4]{\includegraphics[width=2.5in]{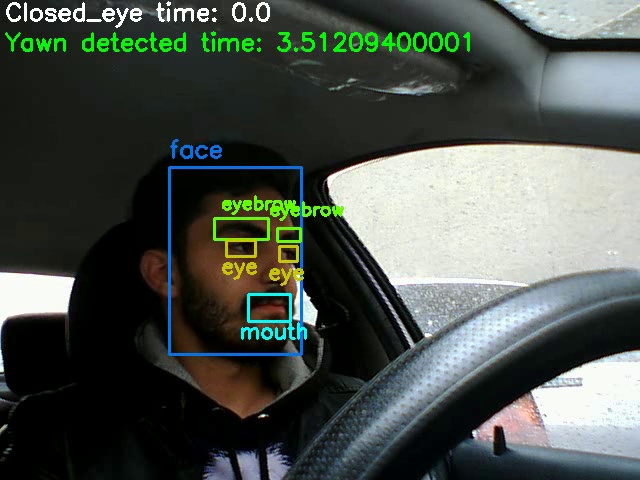}
		\label{b9}}
	\caption{CNN results for different head pose with no glasses. part II}
	\label{fig9_2}
\end{figure}
\begin{figure}
	\centering
	\subfloat[view.1]{\includegraphics[width=2.5in]{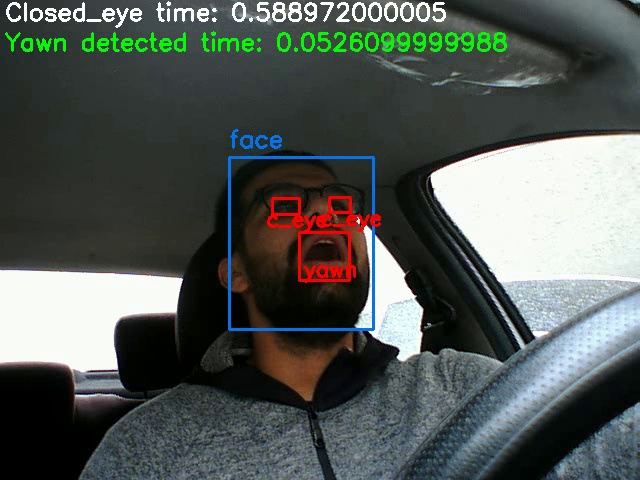}
		\label{a10}}
	\\
	\subfloat[view.2]{\includegraphics[width=2.5in]{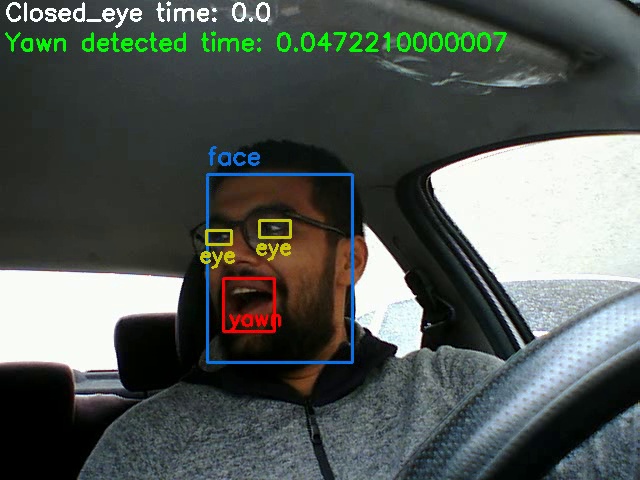}
		\label{c10}} 
	\\
	\subfloat[view.3]{\includegraphics[width=2.5in]{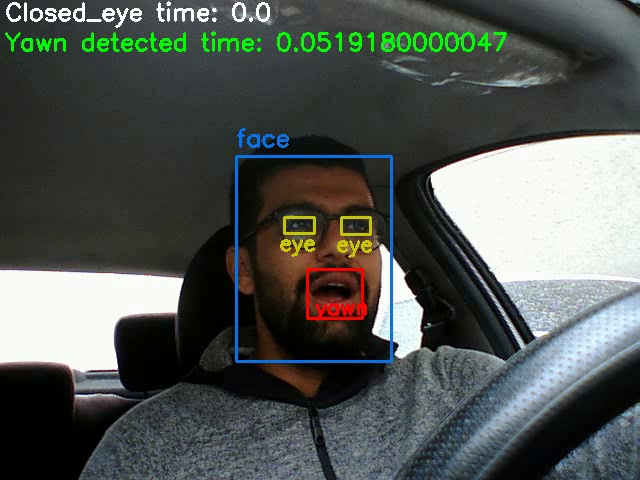}
		\label{e10}}
	\caption{CNN results for different head pose with glasses. part I}
	\label{fig10_1}
\end{figure}
\begin{figure}
	\centering
	\subfloat[view.4]{\includegraphics[width=2.5in]{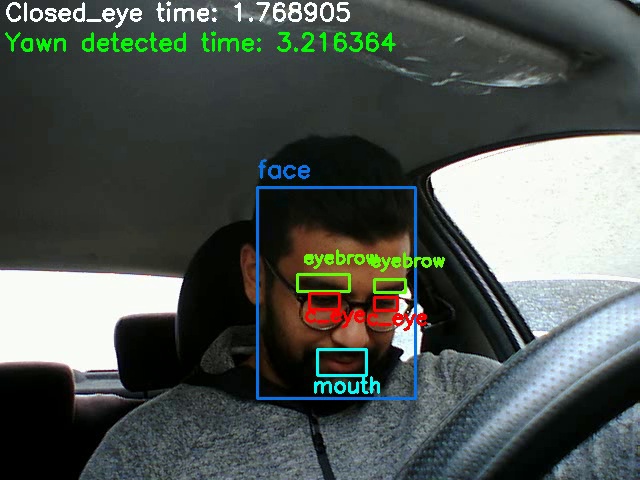}
		\label{b10}}
	\\
	\subfloat[view.5]{\includegraphics[width=2.5in]{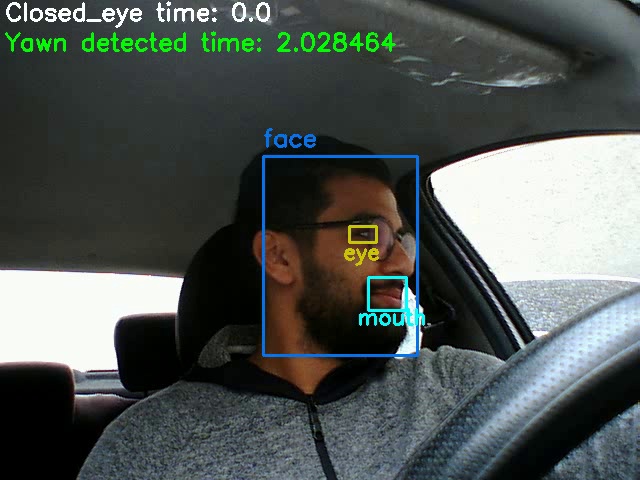}
		\label{d10}} 
	\caption{CNN results for different head pose with glasses. part II}
	\label{fig10_2}
\end{figure}

YOLOv3 is trained for about $4000$ iterations on the training data set; throughout training a batch size of $16$ is exploited along with a momentum of 0.9 and decay of $0.0005$, and finally $0.87$ for mean average precision (mAP) is obtained. Utilizing this result, our network is able to detect every object among the defined labels in almost all head poses due to YOLOv3 depth and rich dataset. Moreover, LSTM is trained for $100$ epochs with batch size of 64 and the $91.7\%$ accuracy is resulted. The LSTM architecture used in this paper is indicated in Fig. \ref{fig11}.      
In the following, results of trained CNN and LSTM are reported.

As it is shown in Fig. \ref{fig9_1} and Fig. \ref{fig9_2} YOLO CNN is capable to detect objects accurately in almost every position. Samples of wearing glasses is depicted in Fig. \ref{fig10_1} and Fig. \ref{fig10_2} indicate the limitation of detecting eyebrows in some positions because of glasses frame. Note that in the reported images besides facial feature labels, two numbers on left-top exist; one is the closed eye time and the other is the time passed after the last yawn detection. These two numbers are going to be used to make the two-dimensional LSTM input.

\begin{figure}[h]
	\centering
	\subfloat[]{\includegraphics[width=2.33in]{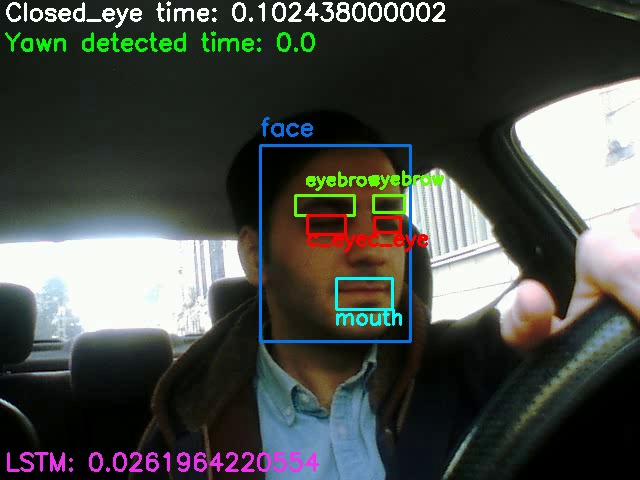}
		\label{L1}}
	\\
	\subfloat[]{\includegraphics[width=2.33in]{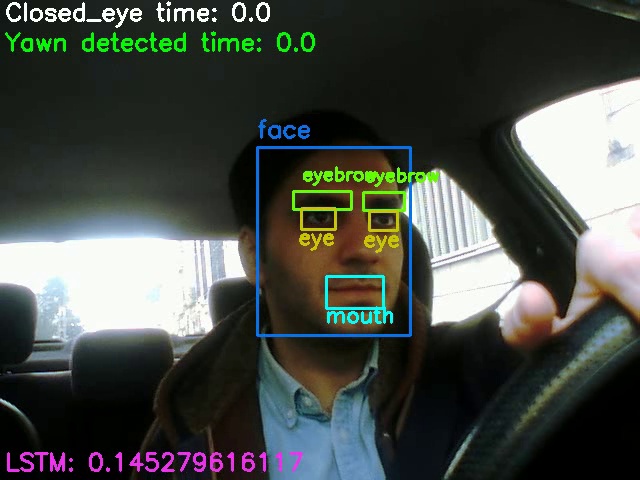}
		\label{L2}} 
	\\
	\subfloat[]{\includegraphics[width=2.33in]{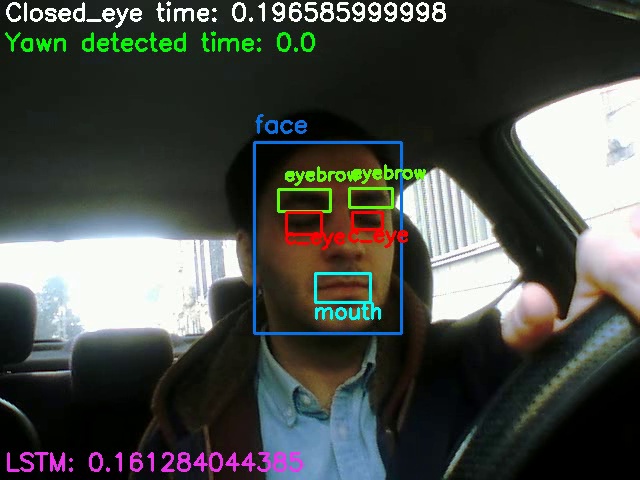}
		\label{L3}}
	\\
	\centering
	\subfloat[]{\includegraphics[width=2.33in]{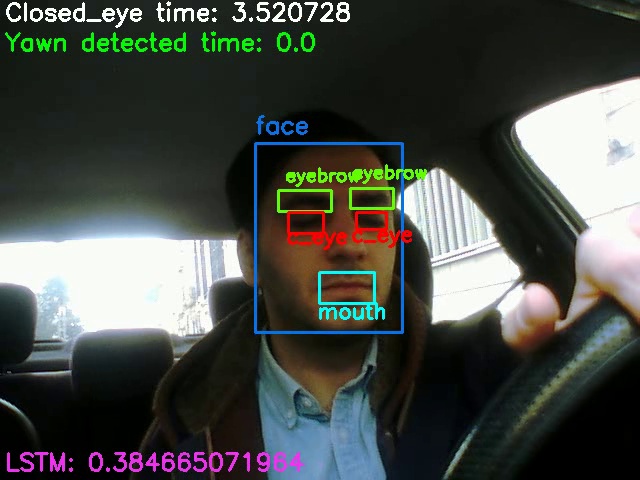}
		\label{L4}}
	\\
	\caption{comparison of LSTM and CNN results. part I}
	\label{fig13_1}
\end{figure}

\begin{figure}[t]
	\centering
	\subfloat[]{\includegraphics[width=2.33in]{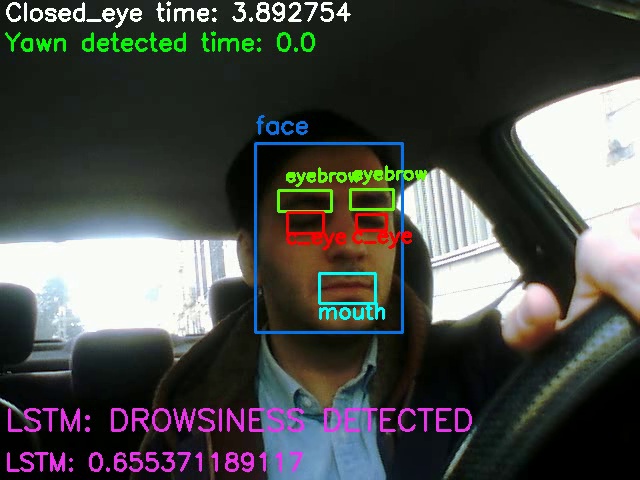}
		\label{L5}}
	\\
	\subfloat[]{\includegraphics[width=2.33in]{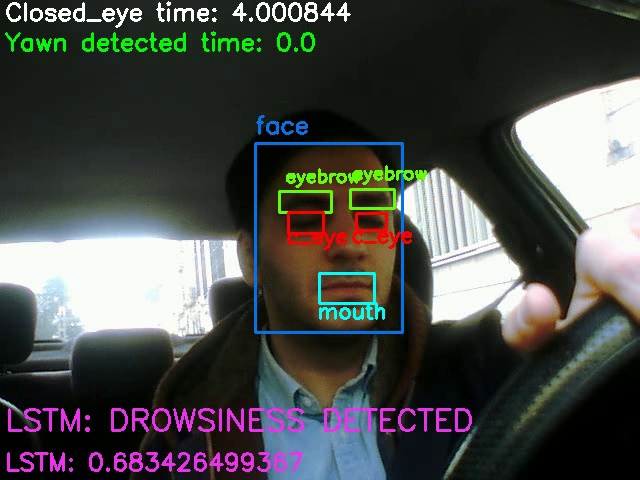}		\label{L6}}
	\\
	\subfloat[]{\includegraphics[width=2.33in]{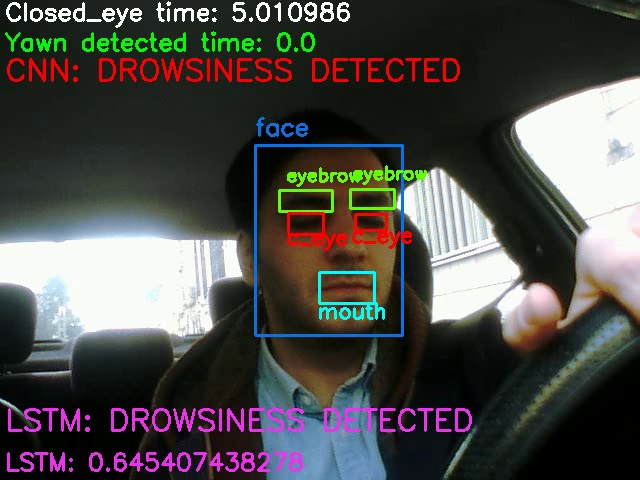}
		\label{L7}}
	\\
	\subfloat[]{\includegraphics[width=2.33in]{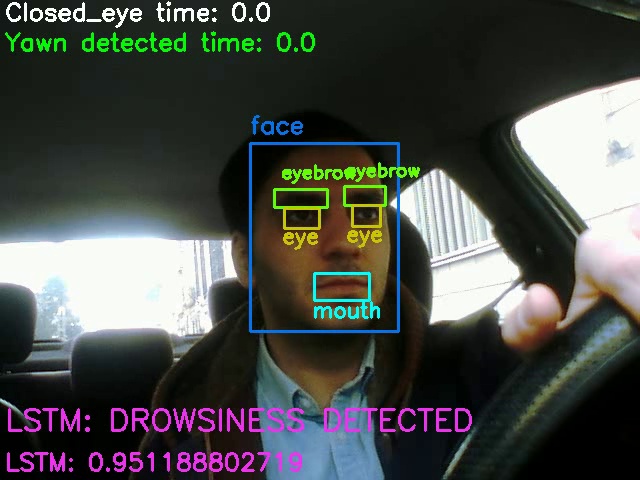}		\label{L8}}
	\caption{comparison of LSTM and CNN results. part II}
	\label{fig13_2}
\end{figure}

Results of the hybrid system of LSTM and CNN are compared with a single CNN based method. In the latter approach mainly two strategies are followed: Since blinking normally takes $300$ to $400$ milliseconds, $5$ seconds can be considered as a threshold for the first strategy which yawning is not considered and just increasing an eye closure duration over the threshold will cause an alarm for fatigue and drowsiness. For second strategy, to take the yawning into account, the threshold is manipulated and decreased to $3$ seconds as a result of yawning detection. To clarify, when yawn is detected, the system warns driver if his/her eyes closure time last more than $3$ seconds. It is remarkable to point out that each time that the yawning is detected; its effect on the threshold remains for $2$ minutes until the sensitivity caused by yawning is eliminated and the strategy comes back to the first state. The decision-making flowchart is depicted in Fig. \ref{fig12}. The CNN approach uses only a threshold and may detect drowsiness too late. Whiles LSTM works in parallel with CNN; uses CNN outputs to learn the driver behavior and predict drowsiness using past sequence of the driver behaviors like blinking slowly.
Considering a single CNN method, each frame can be processed and according to the mentioned strategies the system may warn the driver. This way only the current image frame is checked each time;
thus, the information in the image frame sequence about the
driver behavior is not going to be used in drowsiness detection. The LSTM however, utilizes the data sequence; therefore, in contrast with the single CNN method there is a possibility to detect drowsiness even if the specified threshold is not reached. Moreover, this system can predict drowsiness through incoming data sequences. This big difference is enough to use the hybrid system of LSTM and CNN together instead of a single CNN. To get a better intuition, experimental results are reported in Fig. \ref{fig13_1} and Fig. \ref{fig13_2}, and the comparison is made between the two analyzed approaches. It is notable to insinuate that in each figure at the upper-left corner there are two numbers indicating eye closeness duration and the time passed after the last yawning, respectively. The LSTM output is probability of drowsiness which $50\%$ is considered as threshold to warn the driver.

Fig. \ref{fig13_1} along with Fig. \ref{fig13_2} are displaying a video frame sequence for a driver. Suppose that before Fig. \ref{L1} the driver behavior is normal as the LSTM output is $0.026$. Then the point that $0.102 sec$ duration is measured for eye closeness(as shown in Fig. \ref{L1}), may indicate a starting of a drowsiness process; In Fig. \ref{L2}, although both two times at the upper-left corner are zero, the LSTM output is $0.14$. Furthermore, in Fig. \ref{L3} the eye closeness duration time is $0.19$ thus, the LSTM output is $0.16$ indicating a drowsiness probability of $16\%$. Increase in eye closure duration in Fig. \ref{L4} , \ref{L5} and \ref{L6} is a significant sign of drowsiness as the hybrid system and the single CNN both are warning the driver as in Fig. \ref{L7}. 

The difference between the two mentioned methods is apparent but to show one more excellence of the hybrid system, Fig. \ref{L8} is reported. As it is seen the single CNN right after opened eye detection stops warning the driver because it is just processing the current image frame; Instead, the hybrid system is still warning the driver with even high value of drowsiness probability.
\section{conclusion}
In this article, a method based on a hybrid system of LSTM and CNN was proposed for robust drowsiness detection and prediction. Yolo.v3 was used as the CNN to detect facial features. To train this CNN a new dataset was developed, and transfer learning was used. Furthermore, an LSTM neural network was applied to classify a two-dimensional sequence of data produced by the mentioned CNN. Moreover, recording eye closeness duration and the time passed just after a yawning detected produced a rich dataset for the LSTM training. As an applicable tool to detect and predict drowsiness in real implementation, a multi thread framework was developed to run both CNN and LSTM in parallel to further enhance the performance of the proposed approach. Finally, to show the effectiveness of the presented method, a comparison was made between two cases of using a hybrid system of CNN and LSTM and using a single CNN with two strategies. Finally experiments indicated the robust performance of the proposed approach in drowsiness detection and prediction.
\ifCLASSOPTIONcaptionsoff
\newpage
\fi
\bibliographystyle{IEEEtran}
\bibliography{IEEEabrv,Refarticle}

\end{document}